\documentclass{article}

\usepackage[preprint]{neurips_2026}

\workshoptitle{Sim2Science}

\usepackage{amsmath}
\usepackage{tabularx}
\usepackage[utf8]{inputenc}
\usepackage[T1]{fontenc}
\usepackage{hyperref}
\usepackage{url}
\usepackage{booktabs}
\usepackage{amsfonts}
\usepackage{nicefrac}
\usepackage{microtype}
\usepackage{xcolor}
\usepackage{graphicx}
\usepackage{bm}

\title{\bm{$\pi$}-Net: Empirical Latent Representations for Differentiable Physics Heads in Data-Scarce Systems}

\author{%
  Akshay Dhalpe* \\
  Tampere University \\
  akshay.dhalpe@tuni.fi 
  \And
  Arnab Chattopadhyay \\
  University of Helsinki \\
  arnab.chattopadhyay@helsinki.fi 
  \And
  Shahriar Bakrani Balani \\
  Tampere University \\
  shahriar.bakranibalani@tuni.fi 
  \And
  Sasu Tarkoma \\
  University of Helsinki \\
  sasu.tarkoma@helsinki.fi 
  \And
  Eric Cortanéa  \\
  Tampere University \\
  eric.cortanea@tuni.fi 
}

\begin{document}

\setlength{\abovedisplayskip}{6pt}
\setlength{\belowdisplayskip}{6pt}
\setlength{\abovedisplayshortskip}{0pt}
\setlength{\belowdisplayshortskip}{3pt}

\maketitle

\maketitle

\begin{abstract}
  Developing reliable predictive models for multi-physics systems often forces a compromise between computationally demanding physical simulators and data-hungry, opaque machine learning (ML) algorithms. While incorporating dimensional analysis into directed computation graphs offers a structurally sound alternative to explicitly solving partial differential equations (PDEs), such formulations typically yield underdetermined systems that act as imperfect simulators. We introduce $\pi$-Net, a physics-informed latent variable model that resolves these imperfect simulators. The hybrid  $\pi$-Net architecture  consists of an encoder that maps inputs into physically normalized latent space, which are then decoded into final predictions by a differentiable physics head. Evaluated on a real-world, data-scarce Gas Metal Arc Welding (GMAW) case study, the framework achieves an $R^2$ of 0.864 with only 60 experimental samples. The results demonstrate that $\pi$-Net significantly outperforms purely data-driven baselines in data-scarce regimes while providing causal, process-level explainability.
\end{abstract}

\section{Introduction}
In complex multi-physics domains like advanced manufacturing, predictive models must be physically consistent, interpretable and effective under severe data constraints. Traditional physical simulators rely on complete governing equations, such as  partial and ordinary differential equations (PDEs or ODEs), which are frequently unavailable, partially known, or computationally intensive to solve (\cite{brunton2024promising, chen2018neural}). Conversely, purely data-driven machine learning (ML) models operate as black boxes. They offer no guarantee that predictions satisfy underlying physical dependencies and suffer sharp performance degradation in data-scarce experimental settings (\cite{lee2025toward}). This dichotomy highlights a critical need for hybrid methodologies that can effectively leverage imperfect scientific models.

Physics-informed machine learning (PIML) attempts to bridge this gap, yet most existing approaches still depend on explicit algebraic or differential equations to enforce constraints \cite{karniadakis2021physics, meng2025physics}. The Dimensional Analysis and Conceptual Modelling (DACM) framework \cite{coatanea2016conceptual} provides a powerful alternative by encoding physical phenomena into directed acyclic graphs (DAG) using dimensionless power laws, bypassing the need for explicit PDEs. Despite their flexibility in mapping complex physical interactions, these graph-based representations provide insufficient constraints to uniquely determine all associated parameters \cite{dhalpe2025integrating}. This renders them imperfect simulators that traditionally require biased assumptions to solve.

To address this, we introduce $\pi$-Net, a novel hybrid architecture that reframes these imperfect simulators as physics-informed latent variable models. Specifically, $\pi$-Net utilizes a data-driven ML backbone as an \textbf{encoder} to map physical inputs into a physically normalized and bounded \textbf{latent space} of unresolved dimensionless coefficients ($\boldsymbol{\pi}$). Unlike abstract, statistically derived latent representations, typical in deep learning, these variables capture real, bounded physical efficiencies. A DACM-derived DAG then acts as a fixed, \textbf{differentiable physics head}, projecting these latent variables to the final target prediction. In this framework, the components mutually resolve their fundamental limitations: the differentiable physics head supplies the causal, domain-grounded structure that pure ML lacks. While the data-driven encoder completes the underdetermined simulator directly from data. Crucially, this integration enables true \textbf{end-to-end learning}, where prediction errors backpropagate explicitly through the physical equations in the DAG to guide the encoder.

Our main contributions are as follows:
\begin{itemize}
    \item \textbf{A novel hybrid architecture (\bm{$\pi$}-Net)} that bypasses governing equations, formulating underdetermined simulators as latent variable models by coupling a data-driven encoder to estimate intermediate physical latent variables within a differentiable physics head.

    \item \textbf{End-to-end learning through an imperfect physical simulator.}
    The DAG directly routes both forward predictions and backward error propagation through fixed physical relationships, learning only the unresolved dimensionless coefficients.

    \item \textbf{Demonstrated high data efficiency}, uniformly outperforming pure data-driven baselines in a data-scarce Gas Metal Arc Welding (GMAW) case study.
    \item \textbf{Process-level explainability} that overcomes the opaque nature of standard ML models by explicitly routing the learned latent representations through a causally  physical graph.
\end{itemize}

\section{Related Work}
\textbf{Physics-informed machine learning.}
Physics-informed machine learning (PIML) traditionally embeds prior knowledge through loss functions or architectural constraints based on explicit governing equations, such as in Physics-Informed Neural Networks (PINN) (\cite{karniadakis2021physics, meng2025physics}). While effective, these methods struggle with complex multi-physics systems where the governing equations are incomplete, partially specified, or computationally prohibitive (\cite{brunton2024promising}). Rather than requiring a complete mathematical description, $\pi$-Net incorporates an incomplete scientific model directly into the forward pass as a differentiable physics head. This retains physical structures while the data-driven encoder infers unresolved quantities from data.

\textbf{Dimensional analysis and machine learning.}
To bypass explicit PDEs, some ML approaches leverage dimensional analysis and the Buckingham-$\Pi$ theorem to enforce scale invariance. Recent methods use neural networks, constrained optimization, or symbolic regression to discover these dimensionless groups directly from data (\cite{oppenheimer2023multi, bakarji2022dimensionally}). However, these approaches use dimensionless groups as inputs to learn the overall input-output mapping, leaving the interactions among individual variables implicit in the learned model. In contrast, $\pi$-Net formulates these unknown coefficients as a physically normalized latent space. By decoding this representation through a differentiable physics head: $\pi$-Net backpropagates errors directly through the physical equations, achieving end-to-end learning while preserving causal tracing of variables.

\textbf{Graph-based physical modelling and imperfect simulators.}
The Dimensional Analysis and Conceptual Modelling (DACM) framework constructs DAG using dimensionally homogeneous power laws (\cite{coatanea2016conceptual}). While these directed graphs elegantly capture multiphysics causality, they function as structural templates rather than fully executable simulators, as dimensional homogeneity alone cannot fully constrain the system (\cite{dhalpe2025integrating}).  $\pi$-Net resolves this by treating the unresolved dimensionless coefficients as bounded latent variables estimated by an ML encoder. Instead of learning the physics from scratch, it leverages this latent variable formulation to complete an underdetermined, structurally specified simulator.

\section{Methodology}

\par\noindent\textbf{The \bm{$\pi$}-Net architecture}
Given process inputs $\mathbf{x}$, $\pi$-Net completes the imperfect simulator as a physics-informed latent variable model. The encoder $g_{\theta}$ maps the inputs into a pre-defined, physically normalized latent space of unresolved dimensionless coefficients $\hat{\boldsymbol{\pi}}$. These predicted coefficients are constrained to domain-specified physical intervals $[\pi_i^{\min}, \pi_i^{\max}]$ via a scaled sigmoid activation. The physical computation graph $\mathcal{G}$ then operates as a fixed, differentiable physics head that projects this latent space to the final prediction $\hat{y} = \mathcal{G}(\mathbf{x}, \hat{\boldsymbol{\pi}})$. Thus, physics is embedded as a strict structural prior rather than a soft loss penalty, actively participating in forward and backward passes.

The model is trained end-to-end on paired observations $\mathcal{D} = \{(\mathbf{x}_i, y_i)\}_{i=1}^{N}$ with prediction errors backpropagating directly through the differentiable physics head to update the encoder's parameters $\theta$. As the DAG relies on multiplicative power-law relationships, standard optimization risks numerical instability and gradient explosion. To ensure stable gradient backpropagation through the fixed graph $\mathcal{G}$ to update $\theta$, both the forward pass and the Mean Squared Error (MSE) objective are computed in logarithmic space: $\mathcal{L}(\theta) = \frac{1}{N} \sum_{i=1}^{N} (\log \hat{y}_i - \log y_i)^2$. To validate this framework, the following sections detail the physics head, dataset, and encoder architectures used for evaluation.

\par\noindent\textbf{The GMAW physics head}\label{sec:gmaw_graph}
To demonstrate the $\pi$-Net framework, we instantiate it on a complex multi-physics case study: Gas Metal Arc Welding (GMAW). The objective is to map a set of controllable input variables: arc voltage $U_a$, contact-tip-to-work distance $L$, wire feed rate $v_f$, and travel speed $v_t$ to the target: the weld bead cross-sectional area $A$. 

As the PDEs or ODEs required by traditional Physics-Informed Neural Network (PINN) are unavailable for this process, the  DACM derived DAG encodes the intermediate physical mechanisms (energy transfer, wire melting, and material deposition) through a sequence of dimensionally homogeneous power-law relationships. The forward computational path is defined by the following sequential transformations:

\begin{align}
    U_a &= \pi_1  U_1 \label{eq:Ua} \\
    q_a &= \pi_2 I_1 U_a \label{eq:qa} \\
    q_r &= \pi_3 \lambda_r L I_1^2 d_w^{-2} \label{eq:qr} \\
    q &= q_a + q_r \label{eq:q_tot} \\
    m_l &= \rho_s^{1/2} d_w v_f^{1/2} q^{1/2} \pi_4^{1/2} H_m^{-1/2} \label{eq:ml} \\
    A &= \pi_5 m_l v_t^{-1} \rho_l^{-1} \label{eq:area}
\end{align}

In this formulation, variables such as electrical current $I_1$, wire diameter $d_w$, and enthalpy $H_m$ are known physical constants or measured intermediate quantities. The edges of the computation graph are strictly defined by the physical exponents, representing fixed causal dependencies following the path $\{U_a, L, v_f, v_t\} \rightarrow \{q_a, q_r\} \rightarrow q \rightarrow m_l \rightarrow A$. 

However, the dimensionless coefficients $\{\pi_1,\pi_2, \pi_3, \pi_4, \pi_5\}$ capture complex multiphysics interaction within the robotic GMAW process. The dimensionless coefficients are defined through the functional representation and variable clustering, with each coefficient representing either a single function or a cluster of related functions. Thus, dependencies are formed between the dimensionless coefficients of consecutive functions because the structural dependencies remain fixed. This graph operates as an imperfect simulator that acts as a differentiable physics head, waiting to be completed by the data-driven encoder. 

\par\noindent\textbf{Data and encoders}
To evaluate our framework, we generated a custom experimental dataset specifically for this study using a robotic Gas Metal Arc Welding (GMAW) process. The four controllable input variables ($\mathbf{x}$) were sampled via a Maximum Projection (MAXPRO) design to ensure space-filling diversity, targeting the physical weld cross-sectional area ($y = A$). The prediction target $y$ is the physical weld cross-sectional area $A$.  The resulting dataset contains $N=60$ experimental runs, partitioned equally into two distinct physical regimes: short-circuit ($30$ samples) and globular ($30$ samples) metal-transfer modes. Such data-scarce conditions are inherent to manufacturing use-cases, where each observation requires physical material deposition, thermal stabilization, and costly precision metrology, making large-scale data acquisition time and resource intensive. To isolate the structural benefits of the physics head from representational capacity, we instantiate the encoder $g_{\theta}$ using three distinct architectures (LASSO, MLP, and FT-Transformer) to evaluate on this constrained dataset. Although the encoder predicts sample-dependent coefficient values, their functional identities and interactions are prescribed by the graph. 

\section{Results}

\subsection{Evaluating $\pi$-Net on Accuracy and Robustness}
To establish a baseline, we compared three $\pi$-Net variants against their purely data-driven counterparts. Notably, a traditional PINN baseline could not be included, as the complete governing differential equations required for standard PDE-based regularization are unavailable for this complex multi-physics phenomenon. While the baseline models predict the final target directly from the inputs, the $\pi$-Net variants utilize the identical architectures strictly as encoders to predict the latent space, which is then routed through the differentiable physics head. As each pair shares identical capacity, any performance gains are strictly attributable to the embedded DAG.
We evaluated the models using stratified cross-validation on the 60-sample dataset to isolate the impact of the physics integration.  Table \ref{tab:overall_performance} shows, the $\pi$-Net variants uniformly outperform their data-driven baselines. The $\pi$-Net FT-Transformer has the highest accuracy ($R^2 = 0.864$, MAPE = $6.93\%$), demonstrating that the embedded causal structure rather than the  backbone complexity is the driver of predictive performance.

To assess robustness, training data was progressively reduced from 80\% (5-fold) to 50\% (2-fold). As data decreased, purely data-driven models degraded sharply; the baseline FT-Transformer's $R^2$ fell to 0.679 at the 50\% split (Table \ref{tab:r2_decreasing}). In contrast, the $\pi$-Net variants maintained strong predictive capability. Notably, the linear $\pi$-Net LASSO outperformed the non-linear FT-Transformer baseline across all data splits, confirming that embedded physical constraints provide the necessary structure to succeed in extreme data-scarce environments where data-hungry ML architectures fail to generalize.

\begin{table}[h]
\centering
\caption{Performance comparison of backbone architectures and their $\pi$-Net variants.}
\label{tab:overall_performance}
\resizebox{0.7\textwidth}{!}{
\begin{tabular}{lcccccc}
\toprule
 & \multicolumn{3}{c}{\textbf{Baseline}} & \multicolumn{3}{c}{\textbf{$\pi$-Net}} \\
\cmidrule(lr){2-4} \cmidrule(lr){5-7}
\textbf{Model} & \textbf{$R^2$} & \textbf{MAPE} & \textbf{RMSE}  & \textbf{$R^2$} & \textbf{MAPE} & \textbf{RMSE}  \\
\midrule
LASSO & 0.742 & 10.69\% & 1.919 & 0.840 & 7.92\% & 1.511 \\
MLP & 0.809 & 7.39\% & 1.652 & 0.851 & 7.14\% & 1.461 \\
FT-Transformer & 0.796 & 8.14\% & 1.708 & 0.864 & 6.93\% & 1.394 \\
\bottomrule
\end{tabular}
}
\end{table}

\begin{table}[h]
\centering
\caption{Effect of training set size on the cross-validation $R^2$ performance of backbone architectures and their $\pi$-Net variants.}
\label{tab:r2_decreasing}
\begin{tabular}{lcccccc}
\toprule
 & \multicolumn{2}{c}{\textbf{LASSO}} & \multicolumn{2}{c}{\textbf{MLP}} & \multicolumn{2}{c}{\textbf{FT-Transformer}} \\
\cmidrule(lr){2-3} \cmidrule(lr){4-5} \cmidrule(lr){6-7}
\textbf{Training split} & \textbf{Baseline} & \textbf{$\pi$-Net} & \textbf{Baseline} & \textbf{$\pi$-Net} & \textbf{Baseline} & \textbf{$\pi$-Net} \\
\midrule
80\% train; 20\% test  & 0.742 & 0.840 & 0.809 & 0.851 & 0.796 & 0.864 \\
75\% train; 25\% test & 0.697 & 0.830 & 0.832 & 0.851 & 0.782 & 0.852 \\
67\% train; 33\% test & 0.712 & 0.833 & 0.799 & 0.848 & 0.770 & 0.840 \\
50\% train; 50\% test & 0.680 & 0.821 & 0.792 & 0.825 & 0.679 & 0.825 \\
\bottomrule
\end{tabular}
\end{table}

\subsection{Process level explainability}

\begin{figure}[h]
    \centering
    \includegraphics[width=0.4\textwidth]{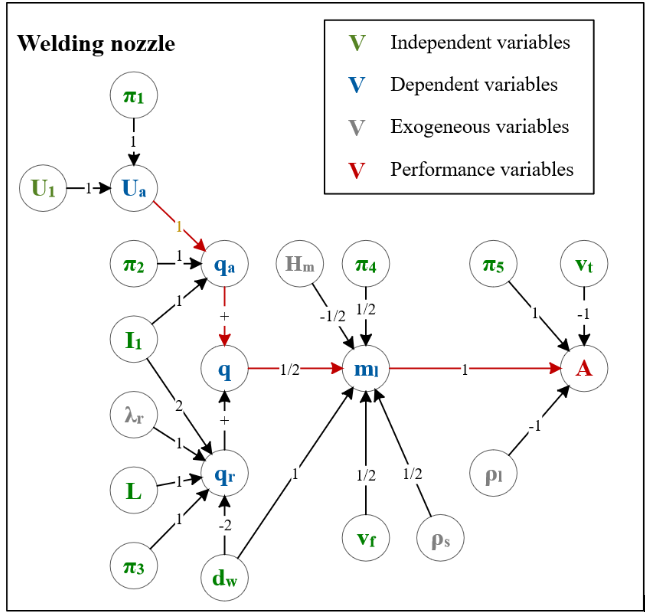}
    \caption{Directed graph of the GMAW process showing the directed propagation path from arc voltage ($U_a$) to weld bead cross-sectional area (A)}
    \label{fig:Explainablity pic}
\end{figure}

As illustrated in Figure \ref{fig:Explainablity pic}, the $\pi$-Net architecture provides structural interpretability by explicitly routing information through a physically meaningful causal graph. For example, the highlighted red pathway demonstrates how a perturbation in the independent variable arc voltage ($U_a$) propagates downstream through intermediate physical states. The directed graph explicitly maps this causal chain: arc voltage dictates arc heat ($q_a$), which contributes to total heat ($q$), driving the wire melting rate ($m_l$), and ultimately determining the final weld bead cross-sectional area ($A$). By maintaining this fixed, dimensionally constrained sequence ($U_a \rightarrow q_a \rightarrow q \rightarrow m_l \rightarrow A$), the framework transcends the opacity of standard black-box ML. It enables domain experts to directly trace how input variations influence final performance targets, providing the process-level explainability required to support reliable engineering decision-making.

\section{Discussion and Future work}

\noindent\textbf{Structural regularization via differentiable physics heads.} Standard data-driven models must implicitly learn physical symmetries and scaling laws from data, a task that requires massive datasets and frequently fails in complex multi-physics environments. Conversely, traditional PINNs require complete partial differential equations, which are often unavailable or computationally prohibitive for such processes. By using a DACM-derived differentiable physics head as a structural prior, $\pi$-Net intrinsically enforces dimensional homogeneity while enabling end-to-end learning. Prediction errors backpropagate explicitly through fixed mathematical relationships to update the encoder. This differentiable prior bounds the hypothesis space to prevent physically impossible predictions and ensure rapid convergence despite severe data scarcity.

 \par\noindent\textbf{Scope and Generalizability.} Despite the graph's regularizing effect, our highly sparse dataset leaves regions of the high-dimensional parameter space underrepresented. Consequently, predictions should be treated with caution when extrapolating out-of-distribution. Furthermore, empirical validation is currently restricted to a single GMAW case study; evaluating the architecture across diverse multi-physics domains remains a critical next step.
 
\par\noindent\textbf{Dimensionless latent spaces and transfer learning.} In standard machine learning, latent spaces are abstract, statistically derived representations lacking physical meaning. In contrast, the $\pi$-Net encoder maps inputs to a latent space of dimensionless coefficients ($\boldsymbol{\pi}$) that represent underlying physical phenomena rather than abstract features or absolute magnitude. These coefficients are coupled through the prescribed functional representation and physical graph, so the model learns graph-consistent latent correction factors rather than independently identifiable physical quantities. Furthermore, the inherent scale-invariance of these physically bounded $\boldsymbol{\pi}$ values might enable zero or few-shot transfer learning across different sized systems.

\par\noindent\textbf{Partial Explainability.} While $\pi$-Net resolves the complete opacity of standard ML, it remains a gray-box architecture. It offers structural interpretability: allowing experts to trace exactly how an estimated latent coefficient causally propagates through intermediate variables. However, while the differentiable physics head is fully transparent, the encoder mapping $g_{\theta}: \mathbf{x} \rightarrow \hat{\boldsymbol{\pi}}$ remains a black box. This decoupled latent architecture represents a pragmatic compromise between the rigid transparency of white-box physical models and the high representational capacity of deep learning.

\par\noindent\textbf{Future work.} $\pi$-Net is viewed as a foundational step toward scalable, structurally constrained representation learning. Future work will extend this via a DACM-GNN architecture, where the DAG governs local thermodynamics and phase transitions while a Graph Neural Network handles spatial diffusion across complex topologies. Ultimately, by leveraging scale-invariant physics inherent to these dimensionless latent variables we aim to study ways to implement zero or few-shot transfer learning, scaling seamlessly from experimental geometries to industrial applications such as Wire Arc Additive Manufacturing (WAAM), battery thermal modeling, and silicon processing.

\begin{table}[htbp]
    \centering
    \renewcommand{\arraystretch}{1.5}
    \begin{tabularx}{\textwidth}{@{} l >{\raggedright\arraybackslash}X >{\raggedright\arraybackslash}X @{}}
        \toprule
        \textbf{Backbone Model} & \textbf{Architecture Configuration} & \textbf{Key Mechanism \& Rationale} \\
        \midrule
        \textbf{LASSO} & 
        Single linear layer mapping the 4 input features to 5 dimensionless coefficient outputs ($\pi_2 - \pi_6$). & 
        Employs an $L_1$ penalty to perform feature selection, identifying which welding parameters exert the greatest influence on each dimensionless group. \\
        
        \textbf{MLP} & 
        \textbf{Input:} 4 neurons \newline
        \textbf{Hidden:} 2 layers (16 neurons each) \newline
        \textbf{Output:} 5 neurons corresponding to the coefficients $\pi_2 - \pi_6$ & 
        Compact design deliberately limits trainable parameters to mitigate overfitting risks on the small experimental dataset ($n=60$). Aligns with standard, robust baselines for GMAW [47,48]. \\
        
        \textbf{FT-Transformer} & 
        \textbf{Tokenization:} Independent linear tokenizer projects each feature to a 64-dimensional embedding with additive learnable positional embeddings. \newline
        \textbf{Encoder:} A CLS token is prepended (forming a 5-token sequence), processed by a 2-layer pre-norm transformer encoder. \newline
        \textbf{Output:} Learned CLS representation is decoded to $\pi_2 - \pi_6$. & 
        Attention mechanism explicitly models the complex, strong interdependencies between welding parameters (e.g., arc voltage and wire feed rate). \\
        \bottomrule
    \end{tabularx}
    \caption{Architectural configuration and structural rationale for the evaluated ML backbones.}
    \label{tab:backbones}
\end{table}

\bibliographystyle{unsrtnat}
\bibliography{bibliography}

\end{document}